\documentclass[letterpaper, 10 pt, conference]{ieeeconf}  % Comment this line out if you need a4paper

\IEEEoverridecommandlockouts                              % This command is only needed if 
\usepackage{epsfig} % for postscript graphics files
\usepackage{times} % assumes new font selection scheme installed
\usepackage{amsmath} % assumes amsmath package installed
\usepackage{amssymb}  % assumes amsmath package installed
\usepackage{graphicx}
\usepackage{hyperref}
\usepackage{xcolor}
\usepackage{multirow}
\usepackage{siunitx}
\usepackage{array}
\usepackage{diagbox}
\usepackage{adjustbox}
\usepackage{booktabs}
\usepackage{todonotes}
\usepackage{pifont}
\usepackage{amssymb}
\usepackage{url}
\usepackage{colortbl}
 \usepackage{subcaption}
\usepackage{pgfplots}
\usepackage{tikz}
\pgfplotsset{width=6cm,compat=1.3}
\usepackage{verbatimbox}

\newcommand{\etal}{\emph{et~al.}}

\newcommand\rebuttall[1]{\textcolor{black}{#1}}

\newcolumntype{P}[1]{>{\centering\arraybackslash}p{#1}}

\renewcommand{\baselinestretch}{0.985}
\newcolumntype{R}[2]{%
    >{\adjustbox{angle=#1,lap=\width-(#2)}\bgroup}%
    l%
    <{\egroup}%
}

\title{\LARGE \bf
Self-Supervised Moving Vehicle Detection from Audio-Visual Cues
}

\author{Jannik Z\"urn and Wolfram Burgard% <-this % stops a space
\thanks{Jannik Z\"urn is with the University of Freiburg, Germany. Wolfram Burgard is with the University of Technology Nuremberg, Germany. Corresponding author: {\tt\small zuern@informatik.uni-freiburg.de}}%
}

\begin{document}

\maketitle
\thispagestyle{empty}
\pagestyle{empty}

%%%%%%%%%%%%%%%%%%%%%%%%%%%%%%%%%%%%%%%%%%%%%%%%%%%%%%%%%%%%%%%%%%%%%%%%%%%%%%%%
\begin{abstract}
Robust detection of moving vehicles is a critical task for any autonomously operating outdoor robot or self-driving vehicle. Most modern approaches for solving this task rely on training image-based detectors using large-scale vehicle detection datasets such as nuScenes or the Waymo Open Dataset. Providing manual annotations is an expensive and laborious exercise that does not scale well in practice. To tackle this problem, we propose a self-supervised approach that leverages audio-visual cues to detect moving vehicles in videos. Our approach employs contrastive learning for localizing vehicles in images from corresponding pairs of images and recorded audio. In extensive experiments carried out with a real-world dataset, we demonstrate that our approach provides accurate detections of moving vehicles and does not require manual annotations. We furthermore show that our model can be used as a teacher to supervise an audio-only detection model. This student model is invariant to illumination changes and thus effectively bridges the domain gap inherent to models leveraging exclusively vision as the predominant modality. 
\end{abstract}

\section{Introduction}
\label{sec:introduction}

% 1 Motivation
Accurate and robust detection of moving vehicles has crucial relevance for autonomous robots operating in outdoor environments~\cite{kummerle2013navigation,kummerle2015autonomous,zurn2020self}. In the context of self-driving cars, other moving vehicles have to be detected accurately even in challenging environmental conditions since knowing their positions and velocities is highly relevant for predicting their future movements and for planning the ego-trajectory. \rebuttall{Also, robots operating in areas reserved for pedestrians such as sidewalks or pedestrian zones, e.g., delivery robots, require precise detections of vehicles in order to assess their planned trajectories and safety clearance with regard to moving vehicles in their vicinity. In this work, we focus on detecting moving vehicles perpendicular to the robot orientation for robots operating in pedestrian areas. Such detections, for example, enable delivery robot applications to more safely cross streets~\cite{kummerle2015autonomous, radwan2020multimodal, radwan2017did} or potentially localize vehicles within pedestrian areas.}

% 2 Current approaches
With the rise of learning-based methods, training image-based vehicle detectors in a supervised fashion has made substantial progress. However, creating large-scale datasets is an expensive and time-consuming task. Previous works have shown that the performance of detectors can break down if they are presented with image data different from samples encountered during training. For outdoor robots, this domain gap may be induced by changes in environmental conditions such as rain, fog, or low illumination during nighttime~\cite{vertens2020heatnet}. While recent approaches in the area of domain adaptation show encouraging results~\cite{vertens2020heatnet,wang2018deep}, most approaches still require a large amount of hand-annotated images in order to not exhibit a large domain gap.

\begin{figure}
\centering
\includegraphics[width=8.5cm]{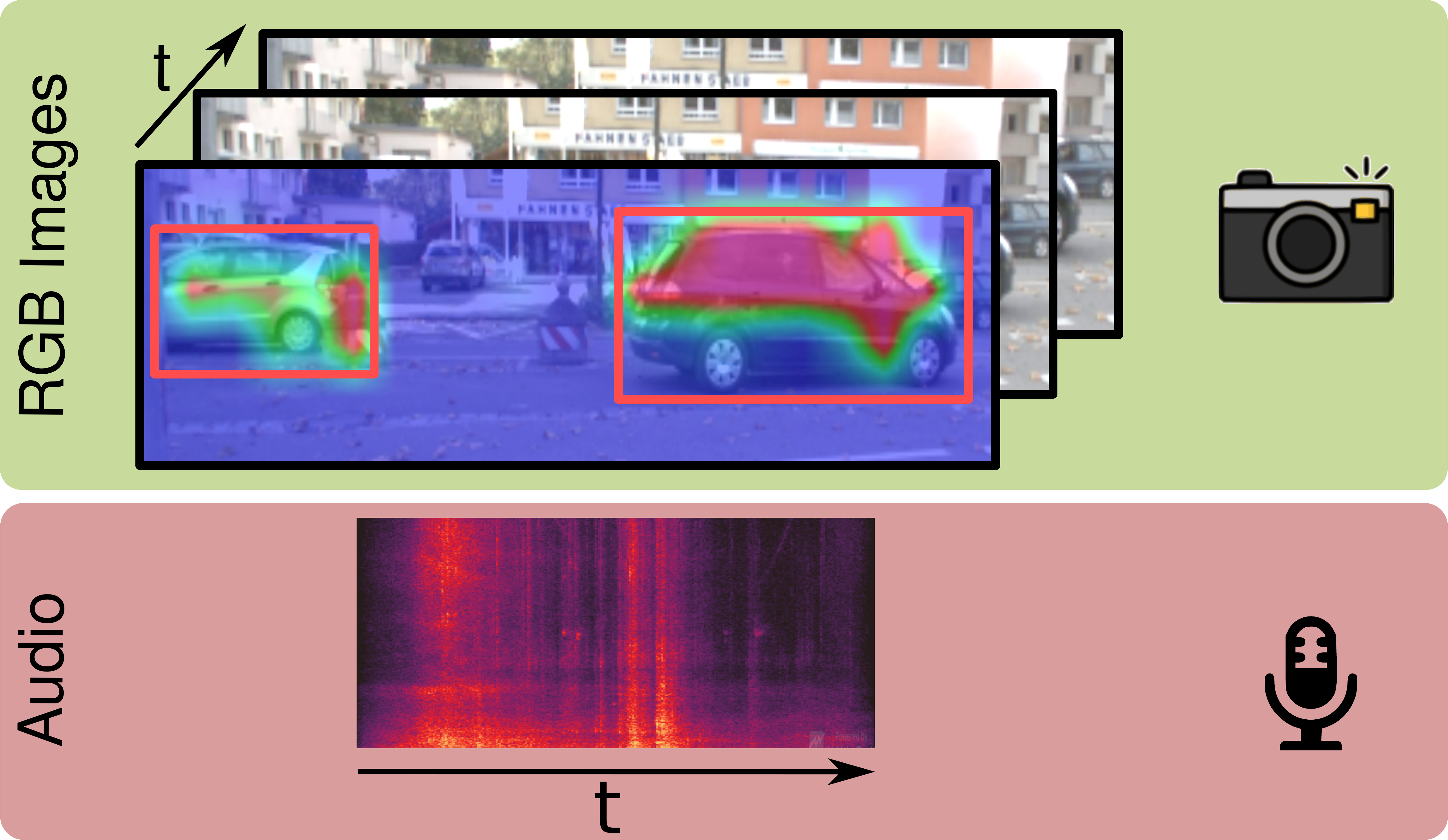}
\caption{We use unlabeled video clips of street scenes recorded with a camera and a multi-channel microphone array and leverage the audio-visual correspondence of features from each modality to train an audio-visual vehicle detection model. The bounding boxes predicted by this model can be used for a downstream student model that does not depend on the visual domain for vehicle detection.}
\label{fig:covergirl}
\end{figure}

Due to the evident domain gap for vision-based detection systems, some works consider the auditory domain instead since it does not exhibit a domain gap between different lighting conditions. The task of sound event localization and detection (SELD) is to localize and detect sound-emitting objects using sound recordings of a multi-channel microphone array. The difference in signal volume and time difference of arrival between the microphones can be exploited to infer the location of a sound-emitting object relative to the microphones~\cite{adavanne2018sound}. However, state-of-the-art learning-based approaches for solving the SELD problem also require hand-annotated labels for training~\cite{baxendale2018audio}, which restricts their applicability in many domains due to the need for large-scale annotated datasets. Self-supervised approaches for audio-visual object detection, in contrast, are able to localize sound-emitting objects in videos without explicit manual annotations for object positions. The majority of these approaches are based on audio-visual co-occurrence of features in both domains~\cite{arandjelovic2018objects, afouras2021self,chen2021localizing}. The main idea in these works is to contrast two random video frames and their corresponding audio segment with each other, leveraging the fact that the chance of randomly sampled frames from different videos containing the same object type is negligible. These works, however, only consider mono or stereo audio instead of leveraging the spatial information in multi-channel audio. Furthermore, they cannot directly be used to predict moving vehicles from videos since vehicles may be present in all videos and at any time. Therefore, the assumption that two randomly sampled videos contain different object types no longer holds true. Lastly, pre-trained object detectors in the visual domain have previously been used as teachers to supervise the detection of sound-emitting objects from the auditory domain~\cite{gan2019self, valverde2021there}. In our work, we do not require a pre-trained teacher model but self-supervise the model using auditory and visual co-occurrence of features.

% 3 Proposition
To overcome these issues, we present a self-supervised approach that leverages audio-visual cues to detect moving vehicles in videos. We demonstrate that no hand-annotated data is required to train this detector and demonstrate its performance on a custom dataset. We furthermore show how this detector can be distilled into a student model which leverages solely the audio modality. In summary, our contributions are as follows:

\begin{itemize}
    \item A novel approach for self-supervised training of an audio-visual model for moving vehicle detection in video clips.
    \item The publicly available \textit{Freiburg Audio-Visual Vehicles} dataset with over 70 minutes of time-synchronized audio and video recordings of vehicles on roads including more than 300 bounding box annotations.
    \item Extensive qualitative and quantitative evaluations and ablation studies of multiple variants of our student and teacher models including investigations on the influence of the number of audio channels and the resistance to audio noise.
\end{itemize}
 
\section{Related Works}
\label{sec:relatedworks}

\subsection{Self-Supervised Audio-Visual Sound Source Localization}

The advancement of Deep Learning enabled a multitude of self-supervised approaches for localizing sounds in recent years. The line of work most relevant for the problem we are considering aims at locating sound sources in unlabeled videos~\cite{arandjelovic2018objects, chen2021localizing, hu2020discriminative, owens2018audio, qian2020multiple}. Arandjelovi\'c and Zisserman~\cite{arandjelovic2018objects} propose a framework for cross-modal self-supervision from video, enabling the localization of sound-emitting objects by correlating the features produced by an audio-specific and image-specific encoding network. Other works consider a triplet loss~\cite{senocak2018learning} for contrasting samples of different classes with each other, while Harwarth \etal~\cite{harwath2018jointly} propose to learn representations that are distributed spatially within and temporally across video frames. These approaches have in common that they exploit the fact that two different videos sampled from a large dataset have a low probability of containing the same objects, while two randomly sampled snippets from the same video have a very high probability of containing the same objects. This can be used to formulate a supervised learning task where the model uses auditory and visual features to highlight regions in videos where sound-emitting objects are visible.

% Why ours is better
In contrast to the works mentioned above, we aim at localizing object instances of a single class, namely moving vehicles, and leverage those detections for a downstream audio-detector model. The assumption that two videos randomly sampled from the dataset have a low likelihood of showing the same object class no longer holds in our application since two different videos may both contain segments with and without vehicles. To circumvent this limitation, we use the audio volume to provide a robust heuristic for classifying image-audio pairs.

\subsection{Audio-Supported Vehicle Detection}

Multiple approaches have been proposed in the last two decades for audio-supported vehicle detection. Chellappa \etal~\cite{chellappa2004vehicle} propose an audio-visual vehicle tracking approach that uses Markov chain Monte Carlo techniques for joint audio-visual tracking. Wang \etal~\cite{wang2012multimodal, wang2012real} propose a multimodal temporal panorama approach to extract and reconstruct moving vehicles from an audio-visual monitoring system for subsequent downstream vehicle classification tasks. Schulz \etal~\cite{schulz2021hearing} leverage Direction-of-Arrival features from a MEMS acoustic array mounted on a vehicle to detect nearby moving vehicles even when they are not in direct line-of-sight. Cross-modal model distillation approaches for detecting vehicles using auditory information were recently investigated intensively \cite{gan2019self, valverde2021there}. Gan \etal~\cite{gan2019self} propose an approach that leverages stereo sounds as input and a cross-modal auditory localization system that can recover the coordinates of moving vehicles in the reference frame from stereo sound and camera meta-data. The authors utilize a pre-trained YOLO-v2 object detector for images to provide the supervision for their approach. Valverde \etal~\cite{valverde2021there} propose a multimodal approach to distill the knowledge of multiple pre-trained teacher models into an audio-only student model.

% Why ours is better
In contrast to previous work, we do not rely on manually annotated datasets to pre-train a model as a noisy label-generator for visual object detection. Instead, we leverage an audio volume-based heuristic to provide supervision to detect moving vehicles in the videos.

\section{Technical Approach}
\label{sec:approach}

\begin{figure*}
\centering
\includegraphics[width=0.95\linewidth]{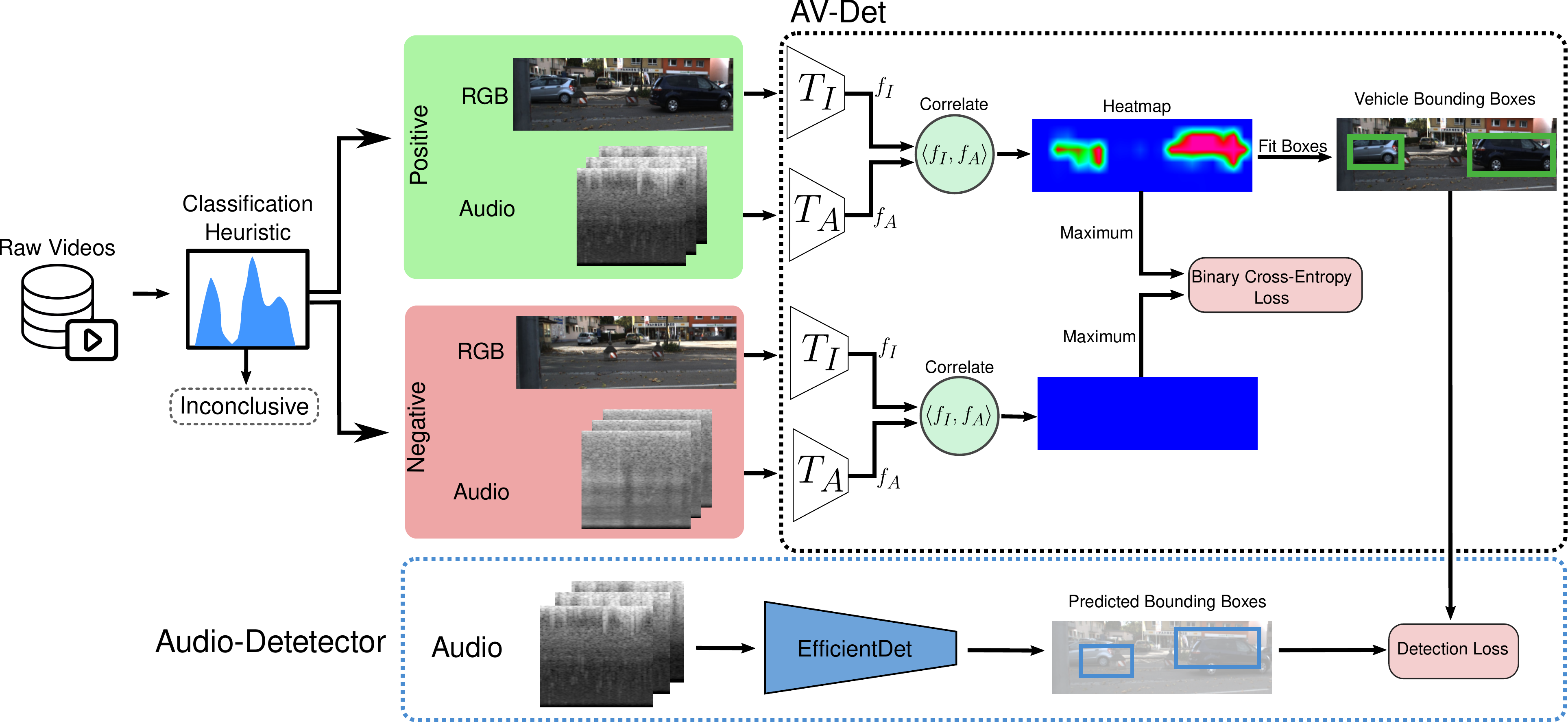}
\caption{We use a volume-based heuristic to classify the videos into positive, negative, and inconclusive image-spectrogram pairs. We subsequently train an audio-visual teacher model, denoted as AV-Det, on positive and negative pairs. We embed both the input image and the stacked spectrograms into a feature space using encoders $T_I$ and $T_A$, producing a heatmap that indicates spatial correspondence between the image features and the audio features. We post-process the heatmap to generate bounding boxes. These bounding boxes can be used to train an optional audio-detector model.}
\label{fig:approach}
\end{figure*}

Pivotal to the motivation of our approach is the observation that sound emitted by passing vehicles and the associated camera images have co-occurring features in their respective domains. Our framework leverages this fact to predict heatmaps for sound-emitting vehicles in the camera images in a self-supervised fashion. It uses these heatmaps to generate bounding boxes for moving vehicles. Subsequently, it uses the bounding boxes as labels for an audio-only model for estimating the direction of arrival (DoA) of the moving vehicles. Figure \ref{fig:approach} illustrates the core components of the approach.

\subsection{Learning to Detect Sound-Emitting Vehicles}

To localize the sound-emitting vehicles in a given image-spectrogram pair, we employ the contrastive learning approach introduced by Arandjelovi\'c  and Zisserman~\cite{arandjelovic2018objects}. We denote an image and its associated audio segment at time step $t$ as the pair $(I_t, A_t)$, where $A_t$ denotes the concatenated spectrograms obtained from the microphone signals temporally centered around the recording time-stamp of the image $I_t$. Following the formulation by Arandjelovi\'c  and Zisserman~\cite{arandjelovic2018objects}, we embed the image $I_t$ in a $C$-dimensional embedding space using a convolutional encoder network $T_I$, producing a feature map $f_I$ of dimensions $H \times W \times C$. Similarly, we embed the audio spectrograms $A_t$ using a convolutional encoder network $T_A$, to produce an audio feature vector $f_A$ with dimensions $1 \times 1 \times C$. We calculate the euclidean distance between each feature vector in the image feature map with the audio feature vector and thus obtain a heatmap $\mathcal{H}$. Formally, the heatmap elements $\mathcal{H}_{mn}$ for image $I$ and multi-channel spectrogram $A$ are defined as

\begin{equation}
\mathcal{H}_{mn} = || T_I(I)_{mn} - T_A(A)||_2^2.
\end{equation}

The localization network $T_I$ is encouraged to reduce the distance between a feature vector from the region in the image containing the sound-emitting object and sound embeddings from audio segments with the same time-stamp while it increases the distance to segments containing no vehicles. This formulation requires knowledge about which sound-emitting objects are visible or audible in each video segment. In previous works, a randomly selected query image-audio pair $(I_q, A_q)$ is contrasted with pairs sampled from different videos since it can be reasonably assumed, due to the large size of the dataset, that the other pairs in the batch contain different object classes than the query pair. With the problem at hand, this assumption is not valid due to the following complications: Each video contains segments both with and without moving vehicles. Thus, the assumption that each video exclusively contains a  single class is incorrect. Therefore, we also cannot assume that two different videos contain different object classes as there are sections in all videos where moving vehicles are present. To overcome this challenge, we instead classify the image-audio pairs in the dataset into three classes: \textit{Positive}, \textit{Negative}, and \textit{Inconclusive}. The \textit{Positive} class entails pairs for which we assume they contain a moving vehicle, while the \textit{Negative} class entails no moving vehicles. The \textit{Inconclusive} class entails pairs for which no clear association can be found. To train our model, we omit the inconclusive pairs as we cannot be certain about their class association. The details of this approach are discussed in Subsection~\ref{subsec:heuristic}. We can now formulate the problem as a binary classification problem similar to Arandjelovic and Zisserman~\cite{arandjelovic2018objects}, where we use a binary cross-entropy loss to learn heatmaps highlighting image regions containing a moving vehicle. Denoting a mini-batch containing positive samples as $B_+$ and negative samples as $B_-$, we arrive at the following per-batch loss:
\begin{eqnarray}
\mathcal{L} &=& - \frac{1}{|B_+| + |B_-|} \left[ \sum_{i=1}^{|B_+|} \log(\max \mathcal{H}_i) \right. \nonumber \\
& & \qquad + \left. \sum_{j=1}^{|B_-|} \log(1 - \max \mathcal{H}_j) \right]
\end{eqnarray}
Minimizing this loss encourages the image encoder network to highlight image regions where the image features are similar to the audio features and dissimilar otherwise.

\subsection{Sample Classification Heuristic}
\label{subsec:heuristic}

\begin{figure}
\vspace{0.2cm}  % Such that figure is not outside margins
    \begin{tikzpicture}
        \begin{axis}[
        xmin=0,
        xmax=600,
        ymin=0.0,
        ymax=0.7,
        height=4cm,
        % xmajorgrids,
        xlabel=Timestep,
        ylabel=$V/V_0$,
        clip=false,
        colormap={BlueToRed}{rgb255(0cm)=(0,0,255);rgb255(0.001cm)=(0,255,0)}
]
        
        \addplot[color=blue,very thick,forget plot]
        % add a plot from table; you select the columns by using the actual name in
        % the .csv file (on top)
        table[x expr=\coordindex+1,y index=0] {figures/vol_data.csv}; 
           
           \draw[thick] (axis cs:\pgfkeysvalueof{/pgfplots/xmin},0.25) -- (axis cs:\pgfkeysvalueof{/pgfplots/xmax},0.25);
           \draw[thick] (axis cs:\pgfkeysvalueof{/pgfplots/xmin},0.38) -- (axis cs:\pgfkeysvalueof{/pgfplots/xmax},0.38);
           
           \node[below right] at (axis cs:620,0.57) {\textit{Positive}};
           \node[below right] at (axis cs:620,0.17) {\textit{Negative}};
           \node[below right] at (axis cs:620,0.37) {\textit{Inconclusive}};
          
        \end{axis}
    \end{tikzpicture}

\caption{Channel-averaged audio volume $V$ over the maximum magnitude $V_0$ for the first $600$ time steps of a recording. We add horizontal bars indicating the upper and lower threshold value for positive, negative, and inconclusive samples.}
\label{fig:audio}
\end{figure}
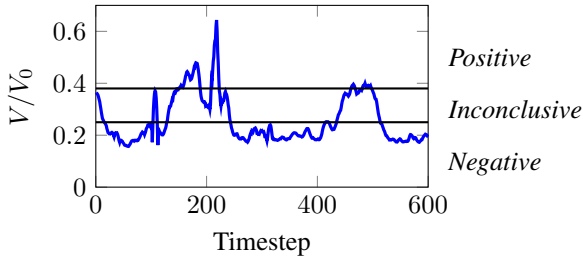

Previously, we glanced over the fact that we require a classifier providing positive and negative samples within a batch. However, in general, we have no access to these labels in the self-supervised learning setting. To circumvent this limitation, we use an audio-volume-based heuristic to classify samples. We leverage two key observations in the data distribution for this classifier: Firstly, frames with low volume typically do not contain any moving vehicles, while louder frames tend to contain moving vehicles as long as the camera is generally directed towards the street and vehicles are located in the camera view frustum. Secondly, we do not require \textit{all} frames to be classified in this fashion. Instead, we require only a subset of all pairs to be classified as long as the number is large enough to prevent over-fitting on the positive and negative samples used for training. Therefore, we label the quietest $N_{q}$ as negative, and the loudest $N_{l}$ pairs from each data collection run as positive pairs. For our experiments, we empirically selected the loudest 15\% and quietest 15\% of samples for all recorded sequences. We found that thresholds between 5\% and 20\% lead to model performances similar to the values reported in Sec. \ref{sec:experiments}. Figure~\ref{fig:audio} illustrates the classification of audio segments based on the audio volume. \rebuttall{The audio volume can potentially also be affected by the ego-robot (motor noise or noise from pavement bumps), however, the system can principally be adjusted to these additional noise sources by adding audio filtering techniques in an optional post-processing step.}

\subsection{Heatmap Conversion to Bounding Boxes}

To be able to quantify the model performance with object detection metrics, we convert the heatmap produced by our model with the following approach: We first clip the heatmap at a threshold of $0.5$. Subsequently, we extract bounding boxes from the clipped heatmap by drawing boxes around connected regions. We do not perform any filtering of or post-processing of the heatmap and also no filtering of the bounding boxes.

\subsection{Audio Student Model}

We adapt our audio student model architecture from the EfficientDet object detection model~\cite{tan2020efficientdet}. The input of the student model are raw channel-concatenated audio spectrograms with a resolution of $512 \times 128$ pixels. The spectrogram corresponds to an audio snippet with a duration of 1 second. We modify the first feature encoding layer to accommodate the varying numbers of input spectrograms. The student model is trained with supervision from the bounding boxes from our AV-Det teacher model as illustrated in Fig. \ref{fig:approach}. We use a focal loss with $\gamma = 2$ as the learning objective for our student model.

\section{Dataset}
\label{sec:dataset}

We collected a real-world video dataset of moving vehicles, the \textit{Freiburg Audio-Visual Vehicles} dataset, which we will make publicly available with the publication of this manuscript. We use an XMOS XUF216 microphone array with 7 microphones in total for audio recording, where six microphones are arranged circularly with a 60-degree angular distance and one microphone is located at the center. The array is mounted horizontally for maximum angular resolution for objects moving in the horizontal plane. To capture the images, we use a FLIR BlackFly S RGB camera and crop the images to a resolution of $400 \times 1200$ pixels. Images are recorded at a fixed frame rate of 5 Hz, while the audio is captured at a sampling rate of 44.1 kHz. The microphone array and the camera are mounted on top of each other with a vertical distance of ca.~$\SI{5}{cm}$.

\begin{figure}
\centering
\includegraphics[width=8.6cm]{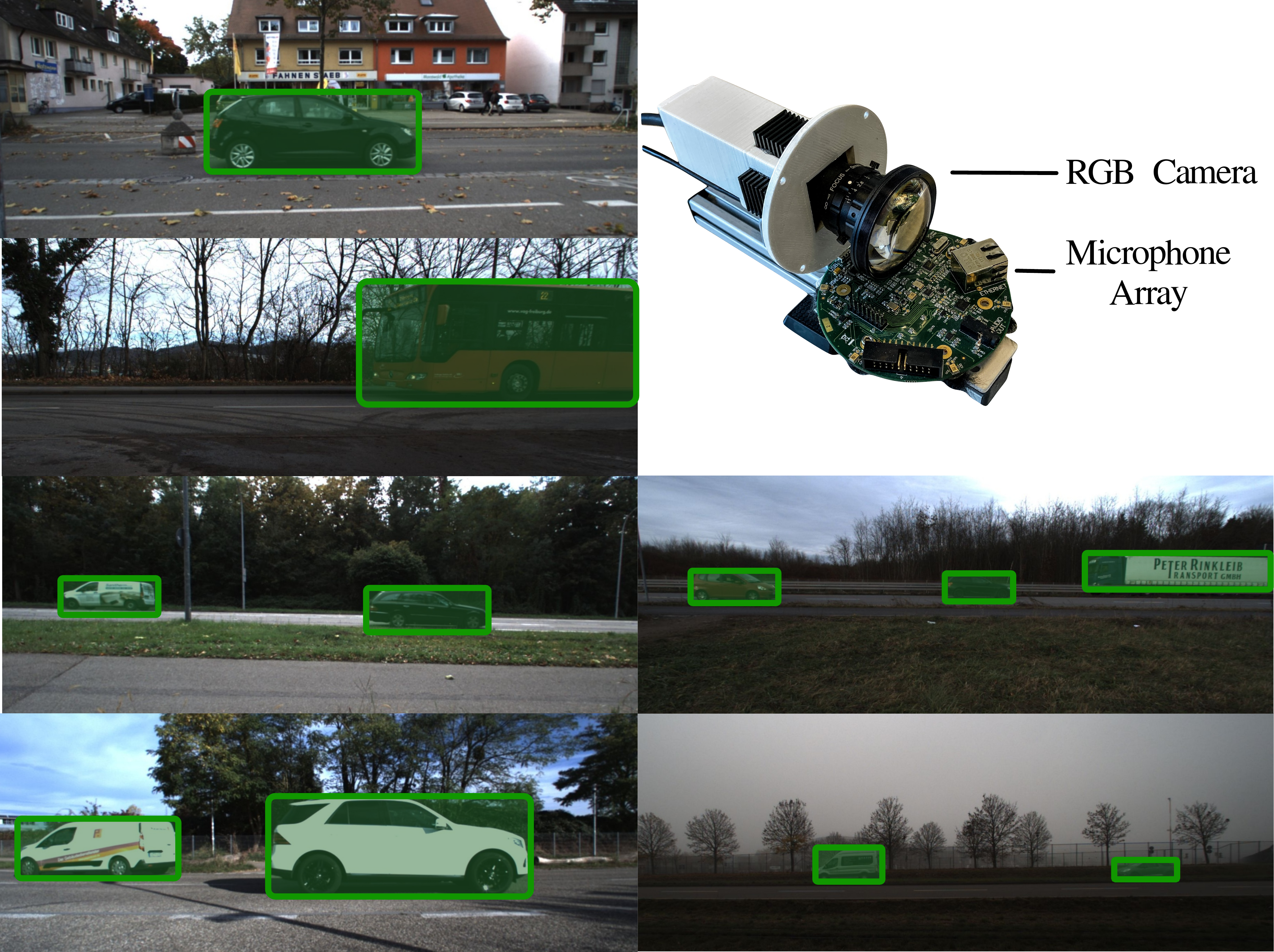}
\caption{Exemplary frames from our Freiburg AudioVisual Vehicles dataset including bounding box annotations for moving vehicles. Scenes include busy suburban streets and rural roads in varying lighting conditions. The data collection sensor configuration is depicted in the top right corner.}
\label{fig:dataset}
\end{figure}

For our dataset, we consider two scenarios: static recording platform and moving recording platform. In the static platform scenario, the recording setup is mounted on a static camera mount. In the dynamic platform scenario, the recording setup is handheld and is moved in a translational fashion (approx. $\SI{15}{cm}$ of positional range along each spatial axis) and rotated with a maximal deviation angle of $\SI{10}{\deg}$. We collected 70 minutes of audio and video footage in nine distinct scenarios with different weather conditions ranging from clear to overcast and foggy. Overall, the dataset contains more than 20k images. The recording environments entail suburban, rural, and industrial scenes. The distance of the camera to the road varies between scenes. To evaluate detection metrics with our approach, we manually annotated more than 300 randomly selected images across all scenes with bounding boxes for moving vehicles. The dataset will be made available at \url{http://av-vehicles.informatik.uni-freiburg.de}.

\section{Experimental Results}
\label{sec:experiments}

We first compare our model with two baseline models (Section~\ref{sec:baselines}). We also perform an ablation study examining the influence of each model component on the detection metrics (Section~\ref{sec:teachereval}) and evaluate the performance of the audio-detector student model (Section~\ref{sec:studenteval}). We furthermore investigate the influence of noise on the heuristic classification accuracy and on the audio-visual detection metrics (Section~\ref{sec:noiseresistance}). To quantify model performance we use the AP value of our only class \textit{Vehicle}. We list AP values for IoU thresholds of $0.1$, $0.2$, and $0.3$, denoted as AP@0.1, AP@0.2, and AP@0.3, respectively. Furthermore, we introduce a bounding box center distance metric, denoted as CD. This distance metric quantifies the average euclidean distance of ground-truth and predicted bounding boxes in each image with an optimal assignment of predicted bounding boxes to ground truth bounding boxes. We introduce the CD metric due to our observation that many predicted bounding boxes are not aligned with the outline of the moving vehicles but instead occupy a subset of the vehicle area or expand over the outline of the vehicles, which leads to box overlaps below the IoU threshold. The CD metric thus helps to quantify the similarity of the box centers instead of the box overlaps. 

\begin{table*}
\centering
\footnotesize
\caption{Performance of models trained and evaluated on the full \textit{Freiburg Audio-Visual Vehicles} dataset and on the dynamic split. Better metric values are indicated with arrows. \rebuttall{The best-performing model results are shown in bold text while the best-performing self-supervised model results are shown in italics.}}
\begin{tabular}{p{3.7cm}|p{1.2cm}p{1.2cm}p{1.2cm}p{1cm}|p{1.2cm}p{1.2cm}p{1.2cm}p{1cm}}
Approach & \multicolumn{4}{c}{Full} & \multicolumn{4}{c}{Dynamic} \\
 & AP@0.1 $\uparrow$	& AP@0.2 $\uparrow$ & AP@0.3 $\uparrow$ & CD $\downarrow$ & AP@0.1 $\uparrow$	& AP@0.2 $\uparrow$ & AP@0.3 $\uparrow$ & CD $\downarrow$ \\
\midrule
Frame Differencing  & $0.3921$ & $0.3411$ & $0.3192$ & $23.22$ & $0.2192$ & $0.0920$ & $0.0470$ & $45.6$ \\
Optical Flow       & $0.4933$ & $0.4890$ & $0.4828$ & $19.15$ & $0.1867$ & $0.1671$ & $0.1633$ & $32.9$ \\
Pre-trained Detector  & $0.6052$  & $0.6047$ & $0.5949$ & $21.14$ & $\mathbf{0.7662}$ & $\mathbf{0.7662}$ & $\mathbf{0.7662}$ & $\mathbf{5.1}$ \\
Pre-trained Detector + Flow  & $0.6634$  & $\mathbf{0.6634}$ & $\mathbf{0.6573}$ & $\mathbf{12.81}$ & $\mathbf{0.7662}$ & $\mathbf{0.7662}$ & $\mathbf{0.7662}$ & $\mathbf{5.1}$ \\
 \midrule
AV-Det 6-channel Oracle       & $0.6054$ & $0.4971$ & $0.3634$ & $19.76$ & $0.6294$ & $0.3418$ & $0.1874$ & $23.2$ \\
 \midrule
AV-Det 1-channel & $0.4169$ & $0.2662$ & $0.1774$ & $25.49$ & $0.1882$ & $0.0966$ & $0.0499$ & $31.1$\\
AV-Det 2-channel & $0.3371$	& $0.2191$ &	$0.1292$ &	$26.50$ &	$0.3724$ &	$0.3083$ &	$0.1233$ &	$36.4$\\
AV-Det 4-channel & $0.4674$	& $0.3587$ &	$0.1901$ &	$21.08$ &	$0.4237$ &	$0.1455$ &	$0.0716$ &	$\mathbf{19.9}$\\
AV-Det 6-channel & $\mathbf{0.7126}$	& $\mathit{0.6195}$ &	$\mathit{0.5575}$ &	$\mathit{16.47}$ &	$\mathit{0.6873}$ &	$\mathit{0.4305}$ &	$\mathit{0.2869}$ &	$33.4$
\label{tab:teacher}
\end{tabular}
\end{table*}

\subsection{Baselines}
\label{sec:baselines}

Due to the lack of prior work for self-supervised audio-visual vehicle detection, we created a flow-based baseline using the RAFT architecture \cite{teed2020raft}. The baseline also does not require any manual annotations and does not rely on pre-trained detectors. The RAFT model was pre-trained on FlyingChairs, FlyingThings3D, Sintel, and KITTI. We use the predicted optical flow to segment moving objects from videos. To obtain bounding boxes from the optical flow model, we first calculate the optical flow between two consecutive frames and threshold the flow field. We finally draw a bounding box around each connected region with a flow value above an empirically found threshold. We also created a baseline based on frame differencing. For this baseline, we threshold the absolute difference between two consecutive frames and extract bounding boxes. We furthermore evaluate the performance of an EfficientDet detector model~\cite{tan2020efficientdet}, which was pre-trained on the MS COCO dataset. We also combine the optical flow baseline with the pre-trained object detector and filter out bounding boxes predicted by the detector if the predicted flow inside a bounding box is below a threshold value. This reduces the risk of the object detector predicting a false-positive static vehicle.

\subsection{Audio-Visual Teacher Evaluation}
\label{sec:teachereval}

In our experiments, we investigate the influence of the number of audio channels. Furthermore, we report the influence of our classification heuristic and compare models trained with labels from the heuristic to a model trained with labels from manual annotations, denoted as \textit{Oracle}. Table~\ref{tab:teacher} lists the results of each model variant. We observe that our best-performing model variant clearly outperforms the frame differencing and vanilla optical flow baselines. All other AV-Det models also mostly outperform the baselines in the dynamic split of our dataset, where many false-positive detections are reported by the frame differencing and optical flow baselines, likely due to the camera motion inducing substantial background pixel differences in consecutive frames. We furthermore report detection metrics from our best-performing AV-Det model that are \rebuttall{mostly on par with the pre-trained EfficientDet + Flow model for the joint static and dynamic split of our dataset and shows the best recorded performance for the AP@0.1 metric. For the dynamic split of the dataset, we observe a deteriorated performance of our model compared to the pretrained model performance}. We note that false-positive detections of the plain pre-trained EfficientDet detector are introduced by the detections of static vehicles in the background. This effect is lessened when the pre-trained detector is combined with the optical flow model to ignore static background vehicle boxes, leading to better detection metrics. In general, we observe that using mono-audio leads to inferior performance compared to the model variants with multiple audio channels.

We also investigate the difference in performance between a model trained with the ground truth classification data and our heuristic. With the heuristic, we report an overall precision of $94.2 \%$ for samples labeled as \textit{Positive} and $77.7 \%$ for samples labeled as \textit{Negative}, respectively. We observe that using the manually annotated ground truth labels does not significantly improve the detection accuracy compared to the heuristic and performs worse than our best-performing AV-Det model trained on classifications obtained from our heuristic. We presume that the heuristic provides less ambiguous training data compared to the manual annotations, which classify an audio-visual sample as \textit{Positive} if at least a part of a moving vehicle is visible in the frame. On one hand, this does not necessarily mean that the vehicle is clearly audible at the same time and on the other hand, a loud vehicle could just have left the camera frustum, leading to a \textit{Negative} audio-video pair that contains sounds characteristic to vehicles but has no correspondence with the image content. Pairs produced with our heuristic suffer to a lesser extent from this inconsistency.

\subsection{Qualitative Results}

\begin{figure*}
\centering
\footnotesize
\setlength{\tabcolsep}{0.0cm}% for the horiz padding
    \begin{tabular}{P{9cm}P{9cm}}
           \includegraphics[width=0.99\linewidth]{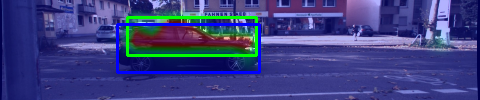} &   \includegraphics[width=0.99\linewidth]{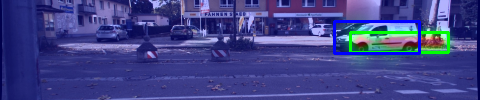} \\
           \includegraphics[width=0.99\linewidth]{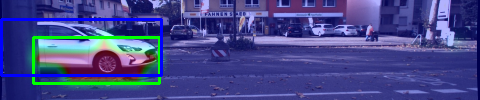} & 
           \includegraphics[width=0.99\linewidth]{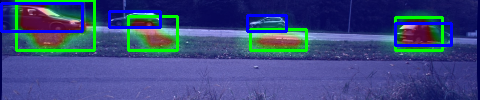} \\
           \includegraphics[width=0.99\linewidth]{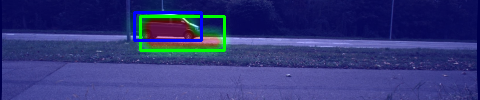} &    \includegraphics[width=0.99\linewidth]{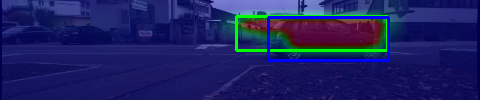} \\
           \includegraphics[width=0.99\linewidth]{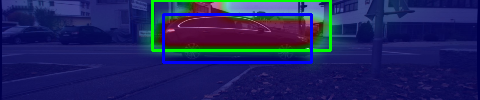} &    \includegraphics[width=0.99\linewidth]{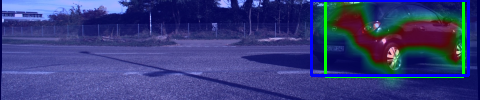} \\
           \includegraphics[width=0.99\linewidth]{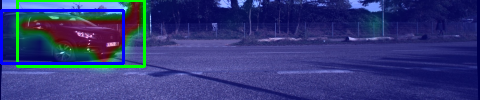} &    \includegraphics[width=0.99\linewidth]{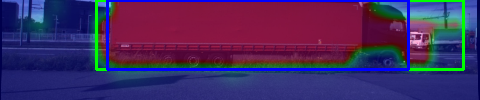} \\

           \noalign{\global\arrayrulewidth=0.5mm}
           \arrayrulecolor{red}\hline  \\

           \includegraphics[width=0.99\linewidth]{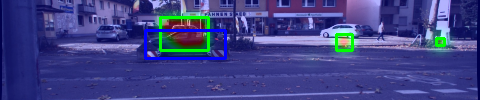} &   \includegraphics[width=0.99\linewidth]{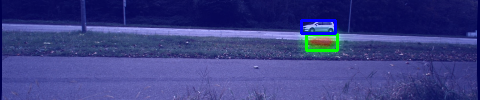} \\
           \includegraphics[width=0.99\linewidth]{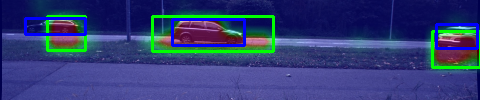} &   \includegraphics[width=0.99\linewidth]{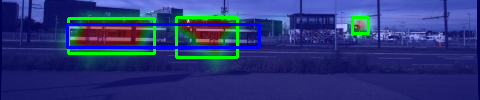} \\

    \end{tabular}
    \caption{Qualitative results of our AV-Det teacher model on the test split of our dataset. We visualize the color-coded learned vehicle heatmap, the estimated bounding boxes in green color, and the ground truth bounding boxes in blue color. The two bottom rows illustrate interesting failure cases of our approach.}
    \label{fig:qualitative} 
\end{figure*}

Figure~\ref{fig:qualitative} illustrates exemplary detections of our best-performing model on our \textit{Freiburg Audio-Visual Vehicles} dataset. We observe that our model generally predicts mostly accurate heatmaps and resulting bounding boxes even in scenes with a large amount of background clutter. Also, scenes with multiple moving vehicles present show high detection accuracy. While the bounding boxes generally overlap well with the ground-truth regions containing moving vehicles, we observe that in some images, the heatmap extends over the actual dimensions of vehicles or is slightly misaligned. We hypothesize that this is due to the imperfect labeling of our volume-based heuristic for assigning image-spectrogram pairs the labels \textit{Vehicle} or \textit{No-Vehicle}, which results in an inaccurate back-propagation of the label signal back to the respective pixel areas in the input images. We further note a failure mode where false-positive detections of vehicles are induced by incorrect highlights in the learned heatmaps. This effect may be caused by background movement correlating with the presence of moving vehicles in the foreground, producing an incorrect association of auditory and visual features. Moreover, the approach may fail to correlate visual and audio features correctly when sound-emitting objects are located outside the camera frustum since the approach assumes such objects to be observable in both domains simultaneously. In our experiments, however, we found the moving vehicles to be the dominant sound source despite audible ambient noise.

\subsection{Audio-Detector Student Model Evaluation}
\label{sec:studenteval}

We evaluate the detection performance of the audio-detector student model in Table~\ref{tab:student}. We run experiments using 1, 2, 4, and 6-channel audio. We also evaluate a student model trained on RGB images instead of audio spectrograms. All audio-detector student models perform worse than the AV-Det teacher model but exhibit an overall decent performance with AP@0.1 values close to $0.4$. In contrast to the AV-Det model, we observe no strong correlation between the number of audio channels and the detection metrics. The 6-channel model version performs similar to the 1-channel version or the other versions regarding the AP@0.1 and AP@0.2 metrics. We hypothesize that the student model architecture, which was adapted from the EfficentDet architecture, can only insufficiently extract additional information from the multiple spectrograms for detecting the position of moving vehicles. This assumption is underlined by the fact that the student model fed with the RGB images instead of the spectrograms performs similar to its teacher model, raising the question of whether further modifications to the student model need to be made to allow for better detection of vehicles solely from multi-channel spectrograms.

\begin{table}
\vspace{0.2cm}  % Such that table is not outside margins
\centering
% \normalsize
\setlength\arrayrulewidth{0.5pt}
\caption{Performance of audio-detector model variants.}
\label{tab:student}
\begin{tabular}{p{3cm}|p{1.2cm}p{1.2cm}p{0.8cm}}
Model & AP@0.1 $\uparrow$ & AP@0.2 $\uparrow$ & CD $\downarrow$ \\
 \midrule
Student 1-channel & $0.3583$ & $\mathbf{0.2756}$ & $40.8$ \\
Student 2-channel & $0.3472$ & $0.2450$ & $\mathbf{36.5}$ \\
Student 4-channel & $0.3799$ & $0.2544$ & $41.5$ \\
Student 6-channel & $\mathbf{0.3833}$ & $0.2636$ & $39.2$ \\
 \midrule
Student RGB Image & $0.6885$ & $0.5727$ & $26.8$
\label{tab:sudent}
\end{tabular}
\end{table}

\subsection{Sensitivity to Noise}
\label{sec:noiseresistance}

\begin{figure}
\begin{tikzpicture}

\pgfplotsset{
    scale only axis,
    scaled x ticks=base 10:0,
    xmin=0, xmax=80
}

\begin{axis}[
  axis y line*=left,
  ymin=0, ymax=1.0,
  xlabel=SNR / dB,
  ylabel=AP@0.1,
  ymajorticks=true,
  ymajorgrids=true,
  yminorgrids=true,
  xmajorgrids=true,
  height=3cm,
]
\addplot[smooth,mark=x,red]
  coordinates{
    (0,0.649)
    (40,0.733)
    (80,0.860)
}; \label{plot_accuracy}
\end{axis}

\begin{axis}[
  height=3cm,
  axis y line*=right,
  axis x line=none,
  ymin=0, ymax=1.0,
  ylabel=Precision,
  legend style={at={(1.0,0.34)},anchor=north east,nodes=right}]
]
\addlegendimage{/pgfplots/refstyle=plot_accuracy}\addlegendentry{Label Heuristic Precision}
\addplot[smooth,mark=*,blue]
  coordinates{
    (0,0.27)
    (40,0.481)
    (80,0.584)
}; \addlegendentry{AV-Det AP@0.1}
\end{axis}

\end{tikzpicture}
  \caption{Precision of our classification heuristic and our model AP@0.1 score on samples corrupted with audio noise.}
  \label{fig:noise} 
\end{figure}
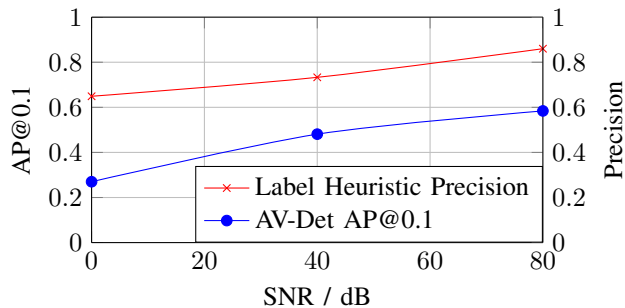

We conduct experiments regarding the noise resistance of our approach. As described in Subsection~\ref{subsec:heuristic}, we use a volume-based heuristic to sort the samples into the classes \textit{Positive}, \textit{Negative}, and \textit{Inconclusive}. One shortcoming of this approach is that objects producing sounds outside the camera frustum can change the prediction of the heuristic due to the change in overall audio signal magnitude. Furthermore, audio noise pollutes the spectrogram and leads to features that are less characteristic of vehicle sounds or the lack thereof. We therefore added varying amounts of white noise to the audio samples in our dataset. We sample the noise from a Gaussian distribution with zero mean and a standard deviation according to the specific Signal-to-Noise Ratio (SNR). We then classify samples using our heuristic and train the AV-Det model with these corrupted samples. Figure~\ref{fig:noise} visualizes the influence of noise on the precision score of our heuristic and the AP@0.1 scores of the AV-Det model. An SNR of $\SI{0}{dB}$ represents noise with the same magnitude as the audio signal while an SNR of $\SI{80}{dB}$ represents a noise signal four orders of magnitude smaller than the signal. We observe that the precision of the heuristic slightly decreases with increasing amounts of noise and arrives at a precision of $0.64$ for an SNR of $\SI{0}{dB}$. The AV-Det model precision only slightly decreases for an SNR of $\SI{40}{dB}$ but is noticeably reduced to $0.27$ for an SNR of $\SI{0}{dB}$. We conclude that our approach shows acceptable performance for lower levels of noise, but its performance deteriorates with higher amounts of noise, \rebuttall{where the precision of our sample classification heuristic decreases. Overall, we conclude that our approach can work well for practical applications where ambient noise does not exceed moderate noise levels but clearly shows improved performance when the moving vehicles are the predominant sound source in a given scene.}

\section{Conclusion}
\label{sec:conclusion}

In this work, we presented a self-supervised audio-visual approach for moving vehicle detection. We showed that an audio volume-based heuristic for sample classification in combination with an audio-visual model trained in a self-supervised manner produces accurate detections of moving vehicles in images. We furthermore demonstrated that the number of audio channels greatly influences the model detection performance, where more audio channels lead to improved performance compared to single-channel audio. Finally, we illustrated that the audio-visual teacher model can be distilled into an audio-only student model, bridging the domain gap inherent to models leveraging vision as the predominant modality. \rebuttall{Possible future work includes 360 degree images for larger visual field-of-view, more diverse scenarios in the dataset, and an improved model architecture for the audio student model.}


%%%%%%%%%%%%%%%%%%%%%%%%%%%%%%%%%%%%%%%%%%%%%%%%%%%%%%%%%%%%%%%%%%%%%%%%%%%%%%%%

{\small
\bibliographystyle{ieee_fullname}
\bibliography{root}

\begin{thebibliography}{10}\itemsep=-1pt

\bibitem{adavanne2018sound}
Sharath Adavanne, Archontis Politis, Joonas Nikunen, and Tuomas Virtanen.
\newblock Sound event localization and detection of overlapping sources using
  convolutional recurrent neural networks.
\newblock {\em IEEE Journal of Selected Topics in Signal Processing},
  13(1):34--48, 2018.

\bibitem{afouras2021self}
Triantafyllos Afouras, Yuki~M Asano, Francois Fagan, Andrea Vedaldi, and
  Florian Metze.
\newblock Self-supervised object detection from audio-visual correspondence.
\newblock {\em arXiv preprint arXiv:2104.06401}, 2021.

\bibitem{arandjelovic2018objects}
Relja Arandjelovic and Andrew Zisserman.
\newblock Objects that sound.
\newblock In {\em Proceedings of the European conference on computer vision
  (ECCV)}, pages 435--451, 2018.

\bibitem{baxendale2018audio}
Mark~D Baxendale, Martin~J Pearson, Mokhtar Nibouche, Emanuele~Lindo Secco, and
  Anthony~G Pipe.
\newblock Audio localization for robots using parallel cerebellar models.
\newblock {\em IEEE Robotics and automation letters}, 3(4):3185--3192, 2018.

\bibitem{chellappa2004vehicle}
Rama Chellappa, Gang Qian, and Qinfen Zheng.
\newblock Vehicle detection and tracking using acoustic and video sensors.
\newblock In {\em 2004 IEEE International Conference on Acoustics, Speech, and
  Signal Processing}, volume~3, pages iii--793. IEEE, 2004.

\bibitem{chen2021localizing}
Honglie Chen, Weidi Xie, Triantafyllos Afouras, Arsha Nagrani, Andrea Vedaldi,
  and Andrew Zisserman.
\newblock Localizing visual sounds the hard way.
\newblock In {\em Proceedings of the IEEE/CVF Conference on Computer Vision and
  Pattern Recognition}, pages 16867--16876, 2021.

\bibitem{gan2019self}
Chuang Gan, Hang Zhao, Peihao Chen, David Cox, and Antonio Torralba.
\newblock Self-supervised moving vehicle tracking with stereo sound.
\newblock In {\em Proceedings of the IEEE/CVF International Conference on
  Computer Vision}, pages 7053--7062, 2019.

\bibitem{harwath2018jointly}
David Harwath, Adria Recasens, D{\'\i}dac Sur{\'\i}s, Galen Chuang, Antonio
  Torralba, and James Glass.
\newblock Jointly discovering visual objects and spoken words from raw sensory
  input.
\newblock In {\em Proceedings of the European conference on computer vision
  (ECCV)}, pages 649--665, 2018.

\bibitem{hu2020discriminative}
Di Hu, Rui Qian, Minyue Jiang, Xiao Tan, Shilei Wen, Errui Ding, Weiyao Lin,
  and Dejing Dou.
\newblock Discriminative sounding objects localization via self-supervised
  audiovisual matching.
\newblock {\em Advances in Neural Information Processing Systems}, 33, 2020.

\bibitem{kummerle2013navigation}
Rainer K{\"u}mmerle, Michael Ruhnke, Bastian Steder, Cyrill Stachniss, and
  Wolfram Burgard.
\newblock A navigation system for robots operating in crowded urban
  environments.
\newblock In {\em 2013 IEEE International Conference on Robotics and
  Automation}, pages 3225--3232. IEEE, 2013.

\bibitem{kummerle2015autonomous}
Rainer K{\"u}mmerle, Michael Ruhnke, Bastian Steder, Cyrill Stachniss, and
  Wolfram Burgard.
\newblock Autonomous robot navigation in highly populated pedestrian zones.
\newblock {\em Journal of Field Robotics}, 32(4):565--589, 2015.

\bibitem{owens2018audio}
Andrew Owens and Alexei~A Efros.
\newblock Audio-visual scene analysis with self-supervised multisensory
  features.
\newblock In {\em Proceedings of the European Conference on Computer Vision
  (ECCV)}, pages 631--648, 2018.

\bibitem{qian2020multiple}
Rui Qian, Di Hu, Heinrich Dinkel, Mengyue Wu, Ning Xu, and Weiyao Lin.
\newblock Multiple sound sources localization from coarse to fine.
\newblock In {\em Computer Vision--ECCV 2020: 16th European Conference,
  Glasgow, UK, August 23--28, 2020, Proceedings, Part XX 16}, pages 292--308.
  Springer, 2020.

\bibitem{radwan2020multimodal}
Noha Radwan, Wolfram Burgard, and Abhinav Valada.
\newblock Multimodal interaction-aware motion prediction for autonomous street
  crossing.
\newblock {\em The International Journal of Robotics Research},
  39(13):1567--1598, 2020.

\bibitem{radwan2017did}
Noha Radwan, Wera Winterhalter, Christian Dornhege, and Wolfram Burgard.
\newblock Why did the robot cross the road?—learning from multi-modal sensor
  data for autonomous road crossing.
\newblock In {\em 2017 IEEE/RSJ International Conference on Intelligent Robots
  and Systems (IROS)}, pages 4737--4742. IEEE, 2017.

\bibitem{schulz2021hearing}
Yannick Schulz, Avinash~Kini Mattar, Thomas~M Hehn, and Julian~FP Kooij.
\newblock Hearing what you cannot see: Acoustic vehicle detection around
  corners.
\newblock {\em IEEE Robotics and Automation Letters}, 6(2):2587--2594, 2021.

\bibitem{senocak2018learning}
Arda Senocak, Tae-Hyun Oh, Junsik Kim, Ming-Hsuan Yang, and In~So Kweon.
\newblock Learning to localize sound source in visual scenes.
\newblock In {\em Proceedings of the IEEE Conference on Computer Vision and
  Pattern Recognition}, pages 4358--4366, 2018.

\bibitem{tan2020efficientdet}
Mingxing Tan, Ruoming Pang, and Quoc~V Le.
\newblock Efficientdet: Scalable and efficient object detection.
\newblock In {\em Proceedings of the IEEE/CVF conference on computer vision and
  pattern recognition}, pages 10781--10790, 2020.

\bibitem{teed2020raft}
Zachary Teed and Jia Deng.
\newblock Raft: Recurrent all-pairs field transforms for optical flow.
\newblock In {\em European conference on computer vision}, pages 402--419.
  Springer, 2020.

\bibitem{valverde2021there}
Francisco~Rivera Valverde, Juana~Valeria Hurtado, and Abhinav Valada.
\newblock There is more than meets the eye: Self-supervised multi-object
  detection and tracking with sound by distilling multimodal knowledge.
\newblock In {\em Proceedings of the IEEE/CVF Conference on Computer Vision and
  Pattern Recognition}, pages 11612--11621, 2021.

\bibitem{vertens2020heatnet}
Johan Vertens, Jannik Z{\"u}rn, and Wolfram Burgard.
\newblock Heatnet: Bridging the day-night domain gap in semantic segmentation
  with thermal images.
\newblock In {\em 2020 IEEE/RSJ International Conference on Intelligent Robots
  and Systems (IROS)}, pages 8461--8468. IEEE, 2020.

\bibitem{wang2018deep}
Mei Wang and Weihong Deng.
\newblock Deep visual domain adaptation: A survey.
\newblock {\em Neurocomputing}, 312:135--153, 2018.

\bibitem{wang2012multimodal}
Tao Wang and Zhigang Zhu.
\newblock Multimodal and multi-task audio-visual vehicle detection and
  classification.
\newblock In {\em 2012 IEEE Ninth International Conference on Advanced Video
  and Signal-Based Surveillance}, pages 440--446. IEEE, 2012.

\bibitem{wang2012real}
Tao Wang and Zhigang Zhu.
\newblock Real time moving vehicle detection and reconstruction for improving
  classification.
\newblock In {\em 2012 IEEE Workshop on the Applications of Computer Vision
  (WACV)}, pages 497--502. IEEE, 2012.

\bibitem{zurn2020self}
Jannik Z{\"u}rn, Wolfram Burgard, and Abhinav Valada.
\newblock Self-supervised visual terrain classification from unsupervised
  acoustic feature learning.
\newblock {\em IEEE Transactions on Robotics}, 37(2):466--481, 2020.

\end{thebibliography}
}

\end{document}